\documentclass{article}

\usepackage{microtype}
\usepackage{graphicx}
\usepackage{subcaption}
\usepackage{booktabs} %

\usepackage{hyperref}

\usepackage[preprint]{icml2026}

\usepackage{amsmath}
\usepackage{amssymb}
\usepackage{mathtools}
\usepackage{amsthm}

\usepackage[capitalize,noabbrev]{cleveref}

\usepackage{amsmath}
\usepackage{booktabs}
\usepackage{multirow}
\usepackage{xcolor}
\usepackage[dvipsnames]{xcolor}
\usepackage{natbib}
\usepackage{graphicx}

\usepackage[dvipsnames]{xcolor}
\newcommand{\best}[1]{\textcolor{Mahogany}{\textbf{#1}}}
\newcommand{\second}[1]{\textcolor{NavyBlue}{\textbf{#1}}}
\newcommand{\refv}[1]{\textcolor{gray}{\textbf{#1}}}

\newcommand{\mname}{
\textsc{Hestia}}
\theoremstyle{plain}
\newtheorem{theorem}{Theorem}[section]

\newtheorem{lemma}[theorem]{Lemma}

\theoremstyle{definition}

\theoremstyle{remark}

\usepackage[textsize=tiny]{todonotes}

\begin{document}

\twocolumn[
  \icmltitle{\textsc{Hestia}: A Hessian-Guided Differentiable Quantization-Aware Training Framework for Extremely Low-Bit LLMs}

  \icmlsetsymbol{equal}{*}

  \begin{icmlauthorlist}
    \icmlauthor{Guoan Wang}{equal,pku_rw}
    \icmlauthor{Feiyu Wang}{equal,pku_cs}
    \icmlauthor{Zongwei Lv}{pku_rw}
    \icmlauthor{Yikun Zong}{pku_cs}
    \icmlauthor{Tong Yang}{pku_cs}
  \end{icmlauthorlist}

  \icmlaffiliation{pku_rw}{School of Software and Microelectronics, Peking University, Beijing, China}
  \icmlaffiliation{pku_cs}{School of Computer Science, Peking University, Beijing, China}

  \icmlcorrespondingauthor{Tong Yang}{yangtong@pku.edu.cn}

  \icmlkeywords{Machine Learning, ICML}

  \vskip 0.3in
]

\printAffiliationsAndNotice{\icmlEqualContribution}

\begin{abstract}
As large language models (LLMs) continue to scale, deployment is increasingly bottlenecked by the memory wall, motivating a shift toward extremely low-bit quantization. However, most quantization-aware training (QAT) methods apply hard rounding and the straight-through estimator (STE) from the beginning of the training, which prematurely discretizes the optimization landscape and induces persistent gradient mismatch between latent weights and quantized weights, hindering effective optimization of quantized models. To address this, we propose \mname{}, a Hessian-guided differentiable QAT framework for extremely low-bit LLMs, which replaces the rigid step function with a temperature-controlled softmax relaxation to maintain gradient flow early in training while progressively hardening quantization. Furthermore, \mname{} leverages a tensor-wise Hessian trace metric as a lightweight curvature signal to drive fine-grained temperature annealing, enabling sensitivity-aware discretization across the model. Evaluations on Llama-3.2 show that \mname{} consistently outperforms existing ternary QAT baselines, yielding average zero-shot improvements of 5.39\% and 4.34\% for the 1B and 3B models. These results indicate that Hessian-guided relaxation effectively recovers representational capacity, establishing a more robust training path for 1.58-bit LLMs. The code is available at \href{https://github.com/hestia2026/Hestia}{https://github.com/hestia2026/Hestia}.
\end{abstract}

\section{Introduction}
Large language models (LLMs) have reshaped the trajectory of artificial intelligence, establishing scaling laws as a reliable path to emergent reasoning and generalization~\cite{achiam2023gpt,guo2025deepseekr1,yang2025qwen3}. However, continued scaling faces constraints from prohibitive computational costs and sustainability concerns. Addressing these challenges has led to a growing emphasis on enhancing representational density, which focuses on optimizing how information is encoded and processed within models. Quantization stands as a critical technique in this paradigm, reducing parameter precision to bridge the gap between massive parameter counts and practical deployment feasibility, while minimizing performance degradation.

\begin{figure*}[t!]
    \centering
    \includegraphics[width=\textwidth]{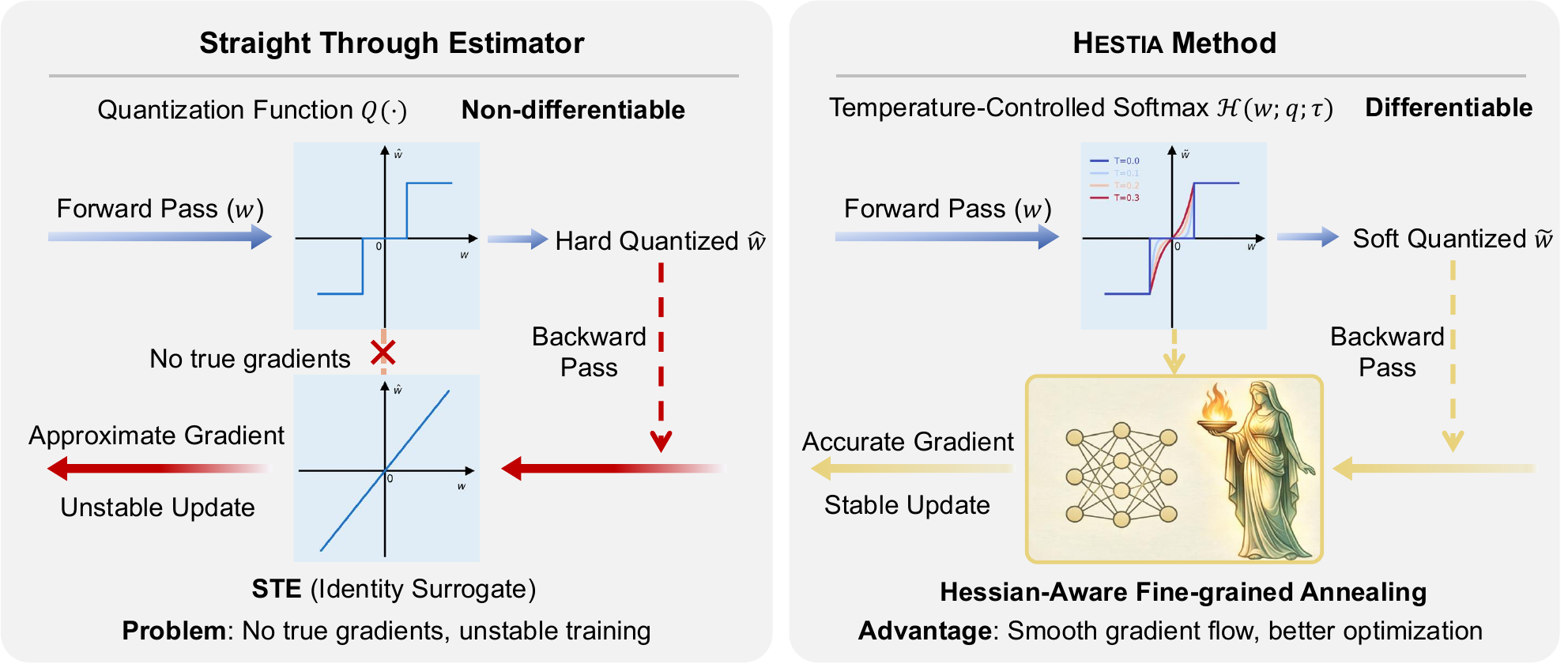}
    \caption{\textbf{Overview of \mname{} (named after Hestia, the Greek goddess of the hearth).} Unlike standard STE (left), which suffers from gradient mismatch due to non-differentiable discrete mapping, \mname{} (right) constructs a differentiable optimization path using a temperature-controlled Softmax surrogate, with a Hessian-trace signal guiding tensor-wise fine-grained temperature annealing.}
    \label{fig:main_framework}
\end{figure*}

Pushing the boundary of representational density leads to extremely low-bit regimes, with ternary architectures representing a significant milestone in this pursuit~\cite{ma2024bitnetb1.58}. By adopting the ternary representation $\{-1,0,1\}$, these architectures enable the replacement of conventional multiply-accumulate (MAC) operations with energy-efficient integer additions. However, the drastic information loss associated with such extreme quantization makes standard Post-Training Quantization (PTQ) challenging, often leading to significant degradation~\cite{xiao2023smoothquant,lin2023awq,chee2023quip,tseng2024quipsharp}. Consequently, maintaining performance at this scale requires Quantization-Aware Training (QAT) to optimize the model parameters within a constrained space. Prevailing ternary QAT approaches simulate discrete weights via rounding and clipping, relying on the Straight-Through Estimator (STE)~\cite{bengio2013ste} for gradient approximation.

Although the use of discrete operators is essential for ensuring consistency between training and inference, STE inherently introduces a mismatch between discrete forward signals and continuous weight updates. Prematurely enforcing such rigid constraints prevents latent weights from crossing discrete boundaries, causing optimization stagnation, a phenomenon frequently characterized as the dead zone problem~\cite{huang2025tequila}. In these dead zones, the updates are insufficient to trigger state transitions, which ultimately traps the model in sub-optimal states before it can learn effective representations. This persistent optimization stagnation prompts a fundamental re-examination of our current methodology:

\begin{center}
\noindent
\colorbox{gray!20}{
    \begin{minipage}{0.95\linewidth}
        \centering
        \textit{Is it appropriate to employ STE-based discrete operators, sacrificing gradient flow during extremely low-bit QAT?}
    \end{minipage}
}
\end{center}

To answer the question, we introduce \textbf{\mname{}}, a Hessian-guided differentiable QAT framework designed to resolve the optimization bottlenecks of extremely low-bit LLMs. To be clarified, this paper mainly focuses on ternary quantization, but \mname{} is also applicable to other quantization schemes. Specifically, \mname{} replaces discrete operators with a temperature-controlled \texttt{Softmax} relaxation. By formulating ternary state assignments as a soft expectation over the discrete set $\{-1,0,1\}$, it provides well-defined, fully differentiable gradients over the latent weight space. This approach effectively mitigates the dead zones that traditionally cause optimization stagnation, allowing weights to move freely during the critical exploration phase.

Crucially, \mname{} exploits the profound heterogeneity of LLM tensors regarding discretization sensitivity. We utilize an offline-calibrated Hessian-trace metric as a lightweight curvature signal to drive an adaptive temperature scaling mechanism~\cite{Dong:2019,tseng2024quipsharp,Tabesh:2025}. Rather than a uniform cosine decay, sensitive tensors with higher curvature maintain higher temperatures for a longer duration. This ensures a more gradual transition to the discrete regime, preserving the representational integrity of critical layers while stabilizing the overall optimization trajectory. Experiments on the Llama-3.2 family~\cite{dubey2024llama3} demonstrate that by reconsidering the training paradigm, \mname{} significantly bridges the performance gap between full-precision and extremely low-bit models. This validates that prioritizing gradient flow over premature hardening is essential for recovering representational capacity in 1.58-bit LLMs. Our main contributions are summarized as follows:
\begin{itemize}  
    \item \textbf{Differentiable LLM QAT Framework:} We introduce a \texttt{Softmax}-based relaxation for extremely low-bit LLMs that ensures well-defined gradients and substantially mitigates optimization dead zones. \item \textbf{Hessian-Aware Temperature Scheduling:} We propose an adaptive mechanism that modulates discretization rates based on second-order curvature to preserve critical model components. \item \textbf{Empirical Validation:} On Llama-3.2-1B and 3B, \mname{} yields average zero-shot improvements of \textbf{5.39\%} and \textbf{4.34\%}, proving that departing from rigid STE is essential for 1.58-bit optimization.
\end{itemize}

\section{Related Work}
The paradigm of extremely low-bit LLMs has been pioneered by BitNet b1.58~\cite{Ma:2024}, which first demonstrated that ternary QAT models can achieve performance competitive with full-precision counterparts. However, this framework typically demands pre-training from scratch to ensure convergence, incurring prohibitive computational costs that hinder widespread adoption. Recent research has focused on optimizing the training recipe itself, guided by quantization scaling laws. For instance, ParetoQ~\cite{Liu:2025} unifies various quantization granularities into a single framework to improve scaling efficiency, while CO-QAT~\cite{Dremov:2025} offers a data-driven path to optimize compute allocation between full-precision pre-training and QAT. However, even with efficient compute allocation, the optimization dynamics remain fundamentally hampered by the gradient mismatch caused by the non-differentiable STE. While PV-Tuning~\cite{Malinovskii:2024} attempts to bypass this via specialized parameterization, other approaches focus on gradient rectification. Tequila~\cite{Huang:2025} identifies the issue as the dead-zone problem, where gradients vanish, and proposes a trapping-free mechanism to reactivate dead weights using dynamic biases, and CAGE~\cite{Tabesh:2025} introduces curvature-aware gradient estimation to reduce variance. Nevertheless, these methods primarily patch the gradient field post-hoc rather than smoothing the underlying optimization landscape.

Other research has pivoted towards differentiable quantization. For instance, DGE~\cite{Wang:2025} introduces a differentiable gradient estimator that approximates gradients of STE but retains hard operations for forward weights, decoupling the forward and backward passes. A more natural solution is continuous relaxation quantization, which approximates the discrete quantization operator with a smooth, fully differentiable function for \textit{both} forward and backward passes. However, few such works exist in the context of LLMs. Classic methods in general neural networks, such as DSQ~\cite{Gong:2019} and DAQ~\cite{Kim:2021}, utilize \texttt{Tanh} sequences or distance-aware \texttt{Softmax} to construct a smooth manifold, gradually increasing the hardness factor to asymptotically approach discrete states. More recently, LLM-specific works have advanced this frontier. For example, LOTION~\cite{Kwun:2025} explicitly smooths the jagged landscape to aid convergence, while others employ stochastic rounding~\cite{Yamazaki:2024} to provide unbiased gradient estimates.
Nevertheless, these approaches typically employ globally uniform annealing schedules or uniform smoothing factors. They overlook the extreme structural heterogeneity of LLMs, where certain layers exhibit much sharper curvature than others. This uniform hardening risks trapping sensitive weights in suboptimal discrete states prematurely, necessitating a rigorous, curvature-informed strategy to dynamically align hardening rates with local landscape complexity.

Second-order curvature, specifically the Hessian trace, has been extensively validated as a robust metric for quantization sensitivity. Pioneering works like HAWQ~\cite{Dong:2019} utilized it to guide spatial mixed-precision allocation, while recent PTQ methods such as QuIP\#~\cite{Tseng:2024} leverage it to minimize reconstruction error. In the realm of QAT, CAGE~\cite{Tabesh:2025} has recently employed curvature to refine gradient estimation, reducing the variance of STE. However, \mname{} takes a distinct geometric path. Since the rigid 1.58-bit constraint prohibits spatial bit-width variation, we repurpose the Hessian trace from a spatial allocator to a temporal scheduler. Instead of varying bit-widths, which violates the 1.58-bit hardware constraint, we use the Hessian trace to modulate the annealing rate of the continuous relaxation, harmonizing the optimization trajectory with the intrinsic curvature of the loss landscape.

\section{Method}
\subsection{Problem Definition}
\label{sec:problem_def}

Consider the optimization of Large Language Models where the weight matrix $\mathbf{W} \in \mathbb{R}^{m \times n}$ is constrained to a ternary set $\mathcal{Q} = \{-1, 0, +1\}$. An element-wise quantization operator $Q: \mathbb{R} \to \mathcal{Q}$ is employed, and the scaling factor $\gamma$ is defined as the mean absolute value stabilized by a small constant $\epsilon$:
\begin{equation}
    \gamma = \frac{1}{N}\sum_{j=1}^{N} |w_j| + \epsilon_{\gamma},
    \label{eq:gamma_def}
\end{equation}
where $N=mn$ denotes the number of elements, $\epsilon_{\gamma}$ is a small constant to avoid division by zero and $w_j \in \mathbf{W}$. In implementation, we treat $\gamma$ as a constant during backpropagation (i.e., $\gamma$ is detached from the computation graph). The dequantized weights $\hat{\mathbf{W}}$ are then obtained via:
\begin{equation}
\scalebox{0.85}{
    $\hat{\mathbf{W}} = Q(\mathbf{W}) = \gamma \cdot \texttt{Clip}\left( \texttt{Round}\left( \frac{\mathbf{W}}{\gamma} \right), -1, +1\right)$.
}
\label{eq:hard_quant}
\end{equation}
Note that $\mathcal{Q}$ denotes ternary \emph{codes}, while $Q(\mathbf{W})$ outputs dequantized weights in $\{-\gamma,0,+\gamma\}$. The optimization objective is to minimize the empirical risk $\mathcal{L}$ over the data distribution $\mathcal{D}$:
\begin{equation}
\scalebox{0.9}{
    $\min_{\mathbf{W}} \mathcal{J}(\mathbf{W})$, \space \text{where} \space $\mathcal{J}(\mathbf{W}) = \mathbb{E}_{x \sim \mathcal{D}} [\mathcal{L}(Q(\mathbf{W}); x)].$
}
\label{eq:discrete_obj}
\end{equation}
Since the gradient of the discrete operator $Q(\cdot)$ vanishes almost everywhere, standard QAT utilizes the STE to approximate the gradient of the objective with respect to the latent weights $\mathbf{W}$. This approach treats the quantization as an identity mapping during the backward pass:
\begin{equation}
    \nabla_{\mathbf{W}} \mathcal{J} = \nabla_{\hat{\mathbf{W}}} \mathcal{L} \cdot \frac{\partial \hat{\mathbf{W}}}{\partial \mathbf{W}} \approx \nabla_{\hat{\mathbf{W}}} \mathcal{L}.
    \label{eq:ste_grad}
\end{equation}

While STE enables backpropagation through the non-differentiable quantizer, it induces a fundamental mismatch between continuous parameter updates and discrete state transitions. For the clipped ternary operator $Q(\cdot)$, the only transition (discontinuity) boundaries occur at
\begin{equation}
    \mathcal{B}=\{-0.5\gamma,\; +0.5\gamma\},
    \label{eq:true_boundaries}
\end{equation}
since \texttt{Clip} removes any output change beyond $|w/\gamma|\ge 1.5$. Let $\Delta w$ denote the latent weight update. We say that a weight $w\in\mathbf{W}$ lies in a dead zone $\Omega_{\mathrm{dead}}$ if this update fails to induce a discrete transition:
\begin{equation}
    \Omega_{\text{dead}} = \left\{ w \in \mathbf{W} \mid Q(w + \Delta w) = Q(w) \right\}.
    \label{eq:dead_zone}
\end{equation}
For $w \in \Omega_{\text{dead}}$, the discrete representation remains unchanged, so the forward computation is insensitive to $\Delta w$ and the objective exhibits a locally flat region despite $\nabla_{\mathbf{W}}\mathcal{J}\neq 0$. This results in stagnation of the optimization process and prevents achieving optimal convergence within the ternary regime.

\subsection{Differentiable Quantization}
\label{sec:homotopy}

To address the issue of optimization stagnation identified in Sec.~\ref{sec:problem_def}, we relax the discrete operator $Q(\cdot)$ into a differentiable surrogate $\mathcal{H}(\cdot; \cdot)$ based on a gradual transition from continuous to discrete states. Specifically, for any latent weight $w \in \mathbf{W}$, we model the assignment to a ternary state $q \in \mathcal{Q}$ through a temperature $\tau$-controlled \texttt{Softmax} kernel $\pi_\tau(q\mid w)$, applied over the normalized distance:
\begin{equation}
\scalebox{0.9}{
    $\pi_\tau(q\mid w)=\texttt{Softmax}(w; q; \tau) = \frac{e^{-(w/\gamma - q)^2 / \tau}}{\sum_{k \in \mathcal{Q}} e^{-(w/\gamma - k)^2 / \tau}}$,
}
\label{eq:softmax_kernel}
\end{equation}

where $\tau$ follows a predefined annealing schedule, controlling the transition from a continuous to a discrete state. The resulting representation $\tilde{w}$ is formulated as the expectation over the state space $\mathcal{Q}$:
\begin{equation}
    \tilde{w} = \mathcal{H}(w; \tau) = \gamma \sum_{q \in \mathcal{Q}} q \cdot \pi_\tau(q\mid w).
    \label{eq:relax_op}
\end{equation}
As the temperature vanishes, the relaxed operator converges to the rigid discrete projection, i.e.,
$\lim_{\tau \rightarrow 0^+}\mathcal{H}(w;\tau)=Q(w)$, so that the model trained in the continuous space is valid for discrete inference. Guided by the annealing schedule, the operator $\mathcal{H}(w; \tau)$ relaxes the rigid ternary state assignment into a soft assignment, bridging the continuous latent space and the discrete ternary regime through the following key properties:
\begin{itemize}
    \item \textit{Landscape Smoothing}: At the beginning of annealing, the operator smooths the optimization landscape, effectively bypassing the dead zones $\Omega_{\text{dead}}$, where the loss surface approximates a linear mapping.
    \item \textit{Gradient Fidelity}: the surrogate provides a well-defined, fully differentiable gradient for $\mathcal{H}(w;\tau)$, so latent updates are faithfully reflected by the differentiable path until assignments become confident as $\tau\to 0^+$.
\end{itemize}
While this differentiable operator guarantees gradient flow, directly projecting to the ternary proxy during early training stages may lead to representational collapse. To mitigate this, we design a \emph{compress stage} and introduce a dynamic pressure parameter $p_t\in[0,1]$. The effective weight used in the forward pass is a time-varying convex combination:
\begin{equation}
\mathbf{W}^{\text{eff}}(t) = (1 - p_t)\cdot \mathbf{W} + p_t \cdot \mathcal{H}(\mathbf{W}; \tau),
\label{eq:weff}
\end{equation}
where 
\begin{equation}
p_t =
\begin{cases}
1, & \rho = 0,\\[2pt]
\min\left(1, \frac{t}{\rho T}\right), & \rho \in (0,1],
\end{cases}
\label{eq:pressure_schedule}
\end{equation}
$t$ is the current step, $T$ is the total number of training steps, and $\rho\in[0,1]$ specifies the fraction of training allocated to the compress stage. The resulting convex interpolation serves as a stabilizing continuation path: at $p_t=0$ the forward pass remains fully in the full-precision regime, while increasing $p_t$ progressively transfers representational responsibility to the quantized surrogate, smoothly steering the model toward ternary representations as $p_t\to 1$. In this way, the schedule separates when quantization is enforced from how discrete assignments harden, yielding a controlled transition that avoids abrupt representational collapse. Consequently, training optimizes a time-varying surrogate objective
\begin{equation}
\tilde{\mathcal{J}}_t(\mathbf{W})=\mathbb{E}_{x\sim\mathcal{D}}\!\left[\mathcal{L}\!\left(\mathbf{W}_{\mathrm{eff}}(t);x\right)\right],
\label{eq:surrogate_obj}
\end{equation}
which recovers the original discrete objective in the limit $p_t\to 1$ and $\tau_t\to 0^+$.

\subsection{Hessian-Aware Annealing Quantization}
\label{sec:hestia_algo}

While the differentiable operator $\mathcal{H}(\cdot; \cdot)$ restores gradient flow, a uniform annealing schedule across all tensors fails to account for the structural heterogeneity of LLMs. From this perspective, we introduce \textbf{\mname{}}, a method that dynamically adjusts the annealing process for each operator based on its sensitivity to quantization, as summarized in Algorithm~\ref{alg:hestia}.

\begin{algorithm}[t]
\caption{\mname{} Algorithm}
\label{alg:hestia}
\begin{algorithmic}[1]
\REQUIRE weights $\mathbf{W}$, total steps $T$, temperature scaling factor $\alpha$, compress ratio $\rho$
\ENSURE dequantized weights $\hat{\mathbf{W}}$
\STATE $\forall i \in \{1, \dots, M\}: s_i \gets \texttt{SensScore}(\mathbf{W}_i)$
\FOR{$t = 0$ \TO $T-1$}
    \STATE batch $B \sim \mathcal{D}$
    \STATE Compute $p_t$ according to Eq.~\eqref{eq:pressure_schedule}.
    \FOR{each operator $i$}
        \STATE $\tau_i \gets \texttt{Schedule}(t, s_i, \alpha)$
        \STATE $\tilde{\mathbf{W}}_i \gets \mathcal{H}(\mathbf{W}_i; \tau_i)$
        \STATE $\mathbf{W}^{\text{eff}}_i \gets (1-p_t)\mathbf{W}_i + p_t \tilde{\mathbf{W}}_i$
    \ENDFOR
    \STATE Assemble $\mathbf{W}^{\text{eff}} \gets \{\mathbf{W}^{\text{eff}}_i\}_{i=1}^{M}$
    \STATE $\mathbf{W} \gets \mathbf{W} - \texttt{Update}\left(\eta; \nabla_{\mathbf{W}} \mathcal{L}(\mathbf{W}^{\text{eff}}; B)\right)$
\ENDFOR
\STATE $\hat{\mathbf{W}} \gets Q(\mathbf{W})$
\end{algorithmic}
\end{algorithm}

\subsubsection{Sensitivity Estimation}
\label{sec:sens_estimation}
In the \mname{} method, we begin by estimating the sensitivity of each operator using the Hessian Trace, which quantifies the curvature and reflects susceptibility to discretization errors. To maintain computational efficiency in LLMs, where the full Hessian is often inaccessible, we approximate this sensitivity metric for each operator $i$ using the Hutch++ algorithm~\cite{meyer2021hutch++}.

Specifically, for operator $i$, let $\mathbf{w}_i=\mathrm{vec}(\mathbf{W}_i)\in\mathbb{R}^{d_i}$ denote the vectorized parameters and
$\mathbf{H}_i=\nabla^2_{\mathbf{w}_i}\mathcal{J}(\mathbf{W})$ the corresponding Hessian.
Hutch++ constructs a low-rank subspace using a sketch matrix $\mathbf{S}_i\in\mathbb{R}^{d_i\times r}$ and an orthonormal basis
$\mathbf{Q}_i$ obtained from the range of $\mathbf{H}_i\mathbf{S}_i$ via QR decomposition.
The trace is then approximated by
\begin{equation}
\scalebox{0.8}{
    $h_i = \text{Tr}(\mathbf{Q}_i^{\top} \mathbf{H}_i \mathbf{Q}_i)
    + \text{Tr}\!\left(\mathbf{G}_i^{\top}(\mathbf{I}-\mathbf{Q}_i\mathbf{Q}_i^{\top}) \mathbf{H}_i (\mathbf{I}-\mathbf{Q}_i\mathbf{Q}_i^{\top})\mathbf{G}_i\right)$,
}
\label{eq:hutchpp_trace}
\end{equation}
where $\mathbf{G}_i$ is a Rademacher random matrix for estimating the residual trace in the orthogonal complement. To address the significant variance in $h_i$ across different tensors, we define a normalized sensitivity score $s_i$ using a standardized \texttt{Sigmoid} transformation:
\begin{equation}
    s_i = \texttt{Sigmoid} \left( \kappa \cdot \frac{\log h_i - \mu_{h}}{\sigma_{h} + \epsilon} \right),
    \label{eq:sens_score}
\end{equation}
where $\mu_{h}$ and $\sigma_{h}$ denote the mean and standard deviation of the log-sensitivity across all linear tensors, and $\kappa$ is a gain factor. Intuitively, a higher $s_i$ reflects greater sensitivity to quantization noise, necessitating softer changes, while a lower $s_i$ allows for harder quantization.

\subsubsection{Temperature Annealing Schedule}
\begin{figure}[t]
    \centering
    \includegraphics[width=\linewidth]{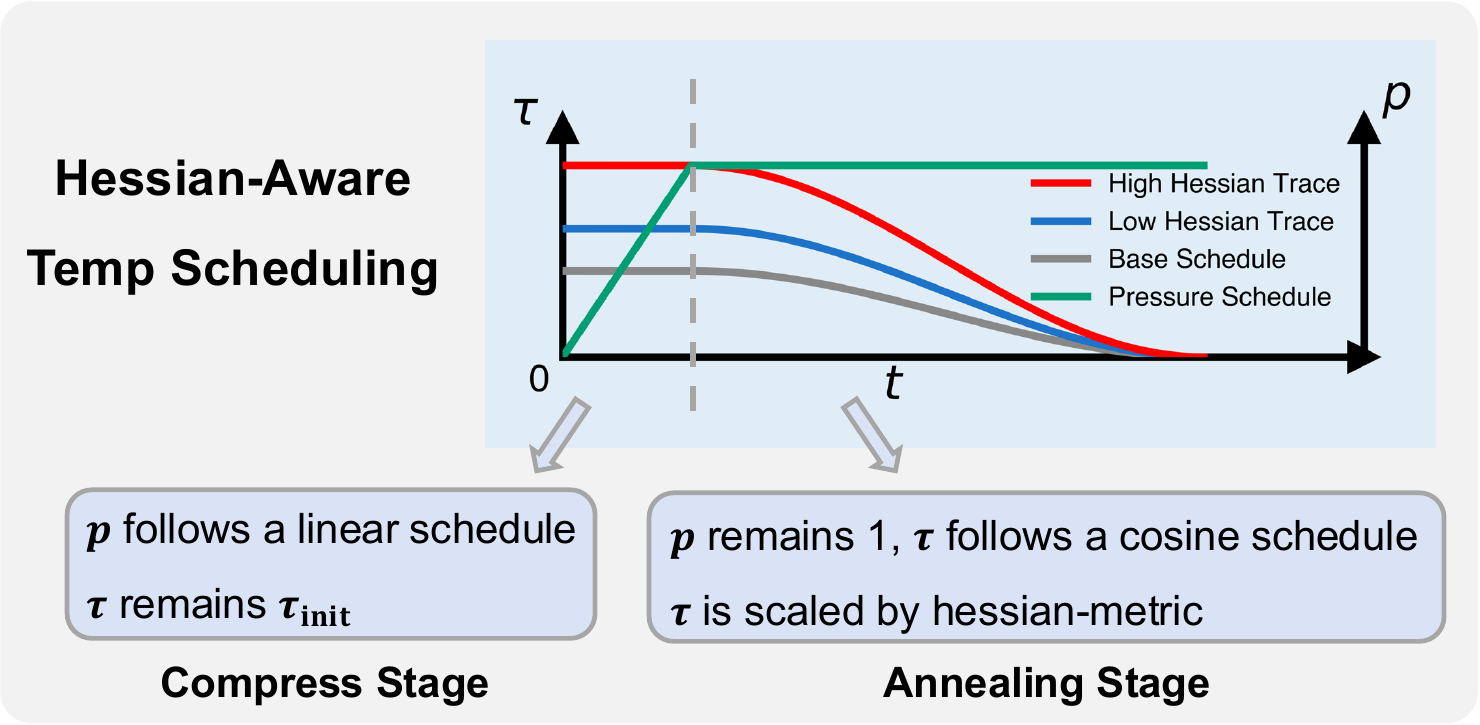}
    \caption{The \mname{} adaptive annealing schedule. In the compress stage, pressure $p$ increases linearly while temperature $\tau$ remains constant. In the annealing stage, $\tau$ decays adaptively to hessian-related metric.}
    \label{fig:hestia_annealing}
\end{figure}

Building upon the sensitivity scores derived in Sec.~\ref{sec:sens_estimation}, we now introduce a temperature annealing schedule that adapts the rate of transition from continuous to discrete states for each operator. We first define a global base temperature schedule $\bar{\tau}(t)$ that follows a standard cosine decay from $\tau_{\text{init}}$ to $0$ over $T$ steps:
\begin{equation}
    \bar{\tau}(t) = \frac{\tau_{\text{init}}}{2} \left[ 1 + \cos \left( \frac{\pi (t-T_{comp})}{T-T_{comp}} \right) \right],
    \label{eq:tau_base}
\end{equation}
where $T_{comp}$ is the time step at which the compression stage ends. To implement \mname{}, we introduce a temperature scaling factor for each operator $i$:
\begin{equation}
    \tau_i(t) = \bar{\tau}(t) \cdot e^{\alpha s_i}.
    \label{eq:tau_i}
\end{equation}
In this formulation, the temperature $\tau_i(t)$ for each tensor is adjusted by the scaling factor $e^{\alpha s_i}$, which is based on the sensitivity score $s_i$ of the tensor. As illustrated in Fig.~\ref{fig:hestia_annealing}, this results in a systematic elevation of the annealing curves relative to the base schedule (green dashed line), where the degree of elevation is monotonic in the sensitivity score and thus in the Hessian-trace-derived curvature proxy.

\section{Theoretical Analysis}
\label{sec:theory}

Building on the relaxed quantizer $\tilde{w}=\mathcal{H}(w;\tau)$ in Eq.~\eqref{eq:relax_op}, we study how the annealed assignment kernel $\pi_\tau(q\mid w)$ in Eq.~\eqref{eq:softmax_kernel} shapes backpropagation.
Specifically, we prove (i) a closed-form Jacobian showing that gradients are modulated by assignment uncertainty, and (ii) a boundary-localization behavior as $\tau\to 0^+$ toward the hard quantization boundaries $\mathcal{B}$ in Eq.~\eqref{eq:true_boundaries}.

Throughout this section, we treat $\gamma$ as a constant during differentiation as in Sec.~\ref{sec:problem_def}. We denote the code mean under $\pi_\tau(\cdot\mid w)$ by
\begin{equation}
\mu_\tau(w)\triangleq \sum_{q\in\mathcal{Q}} q\,\pi_\tau(q\mid w),
\label{eq:mu_tau}
\end{equation}
and the corresponding variance by
\begin{equation}
\mathbb{V}_\tau(w)\triangleq \sum_{q\in\mathcal{Q}} q^2\,\pi_\tau(q\mid w)-\mu_\tau(w)^2 .
\label{eq:var_tau}
\end{equation}
All proofs are deferred to Appendix~\ref{app:theory_proofs}.

\begin{lemma}[Variance-Modulated Jacobian]
\label{lem:variance_gradient}
For $\tilde{w}=\mathcal{H}(w;\tau)$, the Jacobian w.r.t.\ the latent weight satisfies
\begin{equation}
\frac{\partial \mathcal{H}(w;\tau)}{\partial w}
=\frac{2}{\tau}\,\mathbb{V}_\tau(w).
\label{eq:variance_grad}
\end{equation}
\end{lemma}

Lemma~\ref{lem:variance_gradient} implies an uncertainty-gated backward signal:
for any scalar loss $\ell(\tilde{w})$,
\begin{equation}
\frac{\partial \ell(\mathcal{H}(w;\tau))}{\partial w}
=\frac{\partial \ell}{\partial \tilde{w}}\cdot \frac{2}{\tau}\,\mathbb{V}_\tau(w),
\label{eq:uncertainty_gate}
\end{equation}
so gradients are attenuated when $\pi_\tau(\cdot\mid w)$ is confident and amplified when assignments are ambiguous.

\begin{lemma}[Boundary Localization as $\tau\!\to\!0^+$]
\label{lem:boundary_focus}
As $\tau\to 0^+$, the Jacobian $\partial \mathcal{H}(w;\tau)/\partial w$ converges in the sense of
distributions to Dirac measures supported on $\mathcal{B}$:
\begin{equation}
\frac{\partial \mathcal{H}(w;\tau)}{\partial w}
\;\xrightarrow[\tau\to 0^+]{}\;
\gamma \sum_{b\in\mathcal{B}} \delta(w-b).
\label{eq:dirac_limit}
\end{equation}
\end{lemma}

Lemma~\ref{lem:boundary_focus} formalizes a localization effect: annealing progressively concentrates the surrogate Jacobian near decision boundaries, turning smooth continuation into near-discrete refinement.
In practice, optimization operates at finite $\tau>0$, where the nonzero, variance-modulated gradients in
Eq.~\eqref{eq:uncertainty_gate} mitigate dead-zone stagnation.

\begin{table}[t]
\centering
\small
\caption{Zero-shot accuracy on five zero-shot benchmarks.}
\label{tab:main_results}

\begin{subtable}{\linewidth}
\centering
\caption*{\textbf{LLaMA-3.2-1B}}
\vspace{-4pt}
\setlength{\tabcolsep}{4.5pt}
\resizebox{\linewidth}{!}{%
\begin{tabular}{lcccccc}
\toprule
Method & ARC-e & ARC-c & HS & PIQA & WinG & Avg \\
\midrule
BF16 & \refv{0.654} & \refv{0.313} & \refv{0.477} & \refv{0.742} & \refv{0.603} & \refv{0.558} \\
\midrule
LSQ     & 0.376 & 0.177 & 0.258 & 0.574 & 0.506 & 0.378 \\
SEQ     & 0.421 & 0.180 & 0.273 & 0.604 & 0.510 & 0.398 \\
DLT     & 0.424 & 0.174 & 0.256 & 0.563 & 0.513 & 0.386 \\
TWN     & 0.407 & 0.220 & 0.284 & 0.601 & 0.492 & 0.401 \\
AbsMean & 0.603 & 0.259 & 0.360 & 0.683 & 0.541 & 0.489 \\
Tequila & \second{0.645} & \second{0.305} & \second{0.391} & \second{0.710} & \second{0.542} & \second{0.519} \\
INT2  & 0.619 & 0.262 & 0.354 & 0.692 & 0.527 & 0.491 \\
\midrule
\textbf{\mname{}} & \best{0.665} & \best{0.350} & \best{0.418} & \best{0.726} & \best{0.577} & \best{0.547} \\
\bottomrule
\end{tabular}%
}
\end{subtable}

\vspace{6pt}

\begin{subtable}{\linewidth}
\centering
\caption*{\textbf{LLaMA-3.2-3B}}
\vspace{-4pt}
\setlength{\tabcolsep}{4.5pt}
\resizebox{\linewidth}{!}{%
\begin{tabular}{lcccccc}
\toprule
Method & ARC-e & ARC-c & HS & PIQA & WinG & Avg \\
\midrule
BF16 & \refv{0.745} & \refv{0.422} & \refv{0.552} & \refv{0.768} & \refv{0.691} & \refv{0.636} \\
\midrule
LSQ     & 0.431 & 0.200 & 0.294 & 0.599 & 0.522 & 0.409 \\
SEQ     & 0.498 & 0.231 & 0.303 & 0.645 & 0.529 & 0.441 \\
DLT     & 0.361 & 0.161 & 0.260 & 0.572 & 0.496 & 0.370 \\
TWN     & 0.692 & \second{0.351} & 0.462 & 0.734 & 0.586 & 0.565 \\
AbsMean & 0.672 & 0.329 & 0.439 & 0.735 & 0.582 & 0.551 \\
Tequila & \second{0.702} & 0.346 & \second{0.464} & \second{0.739} & \best{0.627} & \second{0.576} \\
INT2  & 0.687 & 0.348 & 0.423 & 0.727 & 0.558 & 0.549 \\
\midrule
\textbf{\mname{}} & \best{0.741} & \best{0.393} & \best{0.491} & \best{0.757} & \second{0.622} & \best{0.601} \\
\bottomrule
\end{tabular}%
}
\end{subtable}

\end{table}

\begin{table*}[t]
\centering
\footnotesize
\setlength{\tabcolsep}{4pt}
\caption{Comparison with other ternary LLMs across model sizes and training token counts (\#Tokens) on five zero-shot benchmarks.
}
\label{tab:ternary_llm_compare_tokens}
\begin{tabular}{l l l c c c c c c c}
\toprule
Model & Size & \#Tokens & ARC-Easy & ARC-Challange & HellaSwag & PIQA & WinoGrande & Avg Score \\
\midrule
LLaMA3.2 & 1B & --   & \refv{0.654} & \refv{0.313} & \refv{0.477} & \refv{0.742} & \refv{0.603} & \refv{0.558} \\
\midrule
DLT & 1B & 10B  & 0.424 & 0.174 & 0.256 & 0.563 & 0.513 & 0.386 \\
SEQ    & 1B & 10B  & 0.421 & 0.180 & 0.273 & 0.604 & 0.510 & 0.398 \\
LLM-QAT           & 1B & 100B & 0.360 & 0.262 & 0.313 & 0.551 & 0.496 & 0.397 \\
BitNet            & 1.3B & 100B & 0.549 & 0.242 & 0.377 & 0.688 & \second{0.558} & 0.483 \\
Spectra           & 1.1B & 100B & 0.563 & 0.246 & \second{0.388} & 0.693 & 0.555 & 0.489 \\
Tequila        & 1B & 10B  & \second{0.645} & \second{0.305} & 0.391 & \second{0.710} & 0.542 & \second{0.519} \\
\midrule
\textbf{\mname{}} & 1B & 10B & \best{0.665} & \best{0.350} & \best{0.418} & \best{0.726} & \best{0.577} & \best{0.547} \\
\midrule
LLaMA3.2 & 3B & --   & \refv{0.745} & \refv{0.422} & \refv{0.552} & \refv{0.768} & \refv{0.691} & \refv{0.636} \\
\midrule
DLT & 3B & 10B  & 0.361 & 0.161 & 0.260 & 0.572 & 0.496 & 0.370 \\
SEQ    & 3B & 10B  & 0.498 & 0.231 & 0.303 & 0.645 & 0.529 & 0.441 \\
LLM-QAT           & 3B & 100B & 0.445 & 0.307 & 0.434 & 0.627 & 0.506 & 0.464 \\
BitNet            & 3B & 100B & 0.614 & 0.283 & 0.429 & 0.715 & 0.593 & 0.527 \\
Spectra           & 3.9B & 100B & 0.660 & 0.319 & \second{0.483} & \second{0.744} & \best{0.631} & 0.567 \\
Tequila        & 3B & 10B  & \second{0.702} & \second{0.346} & 0.464 & 0.739 & \second{0.627} & \second{0.576} \\
\midrule
\textbf{\mname{}} & 3B & 10B & \best{0.741} & \best{0.393} & \best{0.491} & \best{0.757} & 0.622 & \best{0.601} \\
\bottomrule
\end{tabular}
\end{table*}

\begin{table}[t]
\centering
\small
\setlength{\tabcolsep}{4.5pt}
\caption{Fairy2i-related results on \texttt{Qwen2.5-0.5B}.
Ternary-H denotes ternary quantization with \mname{}, and Fairy2i-H denotes Fairy2i quantization with \mname{} for short, respectively.
}
\label{tab:ifairy_qwen05b}
\resizebox{\linewidth}{!}{
\begin{tabular}{lcccccc}
\toprule
Method & ARC-e & ARC-c & HS & PIQA & WinG & Avg \\
\midrule
BF16 & \refv{0.645} & \refv{0.290} & \refv{0.405} & \refv{0.705} & \refv{0.567} & \refv{0.523} \\
\midrule
Ternary & 0.498 & 0.207 & 0.321 & 0.649 & 0.509 & 0.437 \\
Ternary-H & 0.539 & 0.221 & 0.349 & 0.667 & 0.532 & 0.462 \\
\midrule
Fairy2i & 0.540 & 0.238 & 0.357 & 0.678 & 0.540 & 0.471 \\
Fairy2i-H & 0.556 & 0.236 & 0.357 & 0.670 & 0.549 & 0.474 \\
\bottomrule
\end{tabular}
}
\end{table}

\section{Experiments}
\label{sec-exp}

\subsection{Experimental Setup}
\label{sec:exp_setup}

\paragraph{Models.}
We evaluate \mname{} and its competitors on two base LLMs spanning different scales:
\texttt{LLaMA-3.2-1B} and \texttt{LLaMA-3.2-3B}~\cite{dubey2024llama3}.
All models are initialized from their official pretrained checkpoints and then quantization-aware trained under an identical data budget.

\paragraph{Training data.}
QAT is conducted on 10B tokens sampled from Ultra-FineWeb~\cite{wang2025ultrafineweb} with a sequence length of 1024 under a standard causal language modeling objective, matching the data configuration and sampling pipeline used by the baseline methods as described in the Tequila paper~\cite{huang2025tequila}, which ensures a like-for-like comparison.

\paragraph{Quantization setting.}
Our evaluation setting uses weight-only quantization. \mname{} applies ternary quantization to weights with a group-wise quantizer at a group size of 128, while keeping activations in full precision. We quantize all transformer linear layers, including attention and MLP projections, but leave other components in full precision.

\paragraph{Implementation.}
We implement \mname{} using PyTorch.
The training pipeline is designed as a two-stage process with a lightweight pre-computation phase followed by the main quantization-aware training. Prior to QAT, we offline perform the sensitivity estimation via Hutch++ and optimize the initial temperature $\tau_{\mathrm{init}}$ by minimizing the global mean squared error between full-precision and relaxed weights. During the QAT phase, the sensitivity scores are frozen to guide the annealing process, while the model optimization utilizes the AdamW optimizer. \mname{} introduces two coupled schedules: a dynamic pressure parameter $p$ and the Hessian-aware temperature annealing controlled by the strength factor $\alpha$.
The temperature follows a cosine decay schedule, reaching $\tau=0$ at the end of training to ensure a fully hardened quantizer. Comprehensive hyperparameter configurations are detailed in Appendix~\ref{app:imp_details}.

\paragraph{Baselines.}
We benchmark \mname{} against several representative low-bit QAT approaches, each employing distinct optimization philosophies. First, we include hard-quantization QAT baselines trained with STE, using widely adopted ternarization rules such as TWN~\cite{li2016twn}, BitNet b1.58~\cite{ma2024bitnetb1.58} and related AbsMean-based approaches, Spectra~\cite{kaushal2024spectra}, and Tequila~\cite{huang2025tequila}. In addition to the ternary setting, we also evaluate symmetric 2-bit weight quantization as another baseline. Second, we consider parameterized quantizers, which introduce learnable quantization parameters such as scaling factors, represented by LSQ~\cite{esser2020LSQ}, DLT from TernaryLLM~\cite{chen2024ternaryllm}, and SEQ from ParetoQ~\cite{liu2025paretoq}. Finally, we include LLM-QAT~\cite{liu2023llmqat} as a representative of distillation-based quantization frameworks.

\paragraph{Evaluation.}
We evaluate \mname{} using the open-source project \texttt{lm-evaluation-harness}~\cite{eval-harness} on five benchmarks: ARC-Easy, ARC-Challenge~\cite{clark2018arc}, HellaSwag~\cite{zellers2019hellaswag}, PIQA~\cite{bisk2020piqa}, and WinoGrande~\cite{sakaguchi2021winogrande}. We report per-task accuracy and the average across tasks for our method, where
each reported number is the mean over 5 runs adopting the same evaluation protocol as baseline methods to ensure a fair comparison. For competing methods in Table~\ref{tab:main_results} and Table~\ref{tab:ternary_llm_compare_tokens}, we cite
the results from the original papers~\cite{huang2025tequila}.

\begin{table*}[t]
\centering
\small
\setlength{\tabcolsep}{4.5pt}
\caption{Ablations on group size and Hessian-aware scheduling.
\textsc{w/o Hessian} uses a global temperature annealing shared by all layers/tensors.
\textbf{Avg gain (\%)} is computed relative to the \emph{previous row} in the same block (1B or 3B), and is reported only for \textsc{w/o Hessian} at group size 128 and for \textsc{\mname{}}.}
\label{tab:ablation_groupsize_hessian}
\begin{tabular}{l c c c c c c c c c c}
\toprule
Method & Group & Size & ARC-Easy & ARC-Challange & HellaSwag & PIQA & WinoGrande & Avg Score & Avg Gain (\%) \\
\midrule
BF16      & -- & 1B & 0.654 & 0.313 & 0.477 & 0.743 & 0.603 & 0.558 & -- \\
Tequila   & 128 & 1B & 0.645 & 0.305 & 0.391 & 0.710 & 0.542 & 0.519 & -- \\
w/o Hessian      & --   & 1B & 0.659 & 0.322 & 0.410 & 0.724 & 0.571 & 0.537 & -- \\
w/o Hessian      & 128 & 1B & 0.654 & 0.323 & 0.422 & 0.720 & 0.580 & 0.540 & +0.56\% \\
\mname{}           & 128 & 1B & 0.665 & 0.350 & 0.418 & 0.726 & 0.577 & 0.547 & +1.30\% \\
\midrule
BF16      & -- & 3B & 0.745 & 0.422 & 0.552 & 0.768 & 0.691 & 0.636 & -- \\
Tequila   & 128 & 3B & 0.702 & 0.346 & 0.464 & 0.739 & 0.627 & 0.576 & -- \\
w/o Hessian      & --   & 3B & 0.731 & 0.397 & 0.485 & 0.746 & 0.627 & 0.597 & -- \\
w/o Hessian      & 128 & 3B & 0.738 & 0.402 & 0.487 & 0.753 & 0.612 & 0.599 & +0.34\% \\
\mname{}           & 128 & 3B & 0.741 & 0.393 & 0.491 & 0.757 & 0.622 & 0.601 & +0.33\% \\
\bottomrule
\end{tabular}
\end{table*}

\subsection{Main Results}
\label{sec:main_results}
We present our main results in two parts. First, we evaluate \mname{} under weight-only low-bit quantization on \texttt{LLaMA-3.2-1B} and \texttt{LLaMA-3.2-3B} trained on 10B UltraFineWeb tokens, comparing against representative QAT baselines in Table~\ref{tab:main_results}. Second, we further position \mname{} against a broader set of ternary LLM training recipes across different model sizes and token budgets in Table~\ref{tab:ternary_llm_compare_tokens}. Throughout the main-result tables, \textcolor{gray}{gray} numbers indicate the BF16 reference. Among quantized models, the best value in each column is highlighted in \best{red}, and the second-best value is highlighted in \second{blue}. These results demonstrate that \mname{} consistently achieves improvements in zero-shot accuracy across both smaller and larger models, highlighting its effectiveness in optimizing low-bit quantization for LLMs.

\paragraph{\mname{} improves over strong QAT baselines.}

Table~\ref{tab:main_results} reports zero-shot accuracy on five typical benchmarks. W2-Sym denotes symmetric 2-bit weight quantization as an additional reference point. Across both model scales, \mname{} consistently outperforms the state-of-the-art ternary QAT competitors.
On \texttt{LLaMA-3.2-1B}, \mname{} improves the 5-task average from 0.519 (Tequila) to 0.547, with particularly large gains on ARC-Challenge (+4.5 points) and WinoGrande (+3.5 points).
On \texttt{LLaMA-3.2-3B}, \mname{} further raises the average from 0.576 (Tequila) to 0.601, improving all five benchmarks and yielding notable gains on HellaSwag (+2.7 points) and WinoGrande.
These results indicate that \mname{} provides a more favorable optimization trajectory under extreme quantization, translating into stronger zero-shot generalization.
We additionally compare against a symmetric 2-bit weight-only baseline (\textsc{INT2}).
Despite operating in a more constrained ternary setting, \mname{} achieves higher average accuracy than \textsc{INT2} on both 1B and 3B, suggesting that \mname{}'s training dynamics improvements are not merely a byproduct of a higher-precision codebook, but stem from a better soft-to-hard training path.

\paragraph{Comparison with prior ternary LLMs.}
Table~\ref{tab:ternary_llm_compare_tokens} further positions \mname{} against a broad set of ternary LLM training recipes, covering both 10B-token and 100B-token regimes.
Under the same 10B-token budget, \mname{} consistently yields the strongest accuracy among all quantized models for both 1B- and 3B-class backbones, improving the 5-task average from 0.519 to 0.547 on 1B and from 0.576 to 0.601 on 3B over TequilaLLM.
Notably, \mname{} trained with only 10B tokens is competitive with, and often surpasses, several 100B-token ternary LLM baselines (e.g., BitNet and Spectra) on multiple benchmarks.
These results suggest that \mname{} improves the optimization trajectory of extreme quantization through its gradual soft-to-hard schedule, which enables stronger zero-shot generalization without requiring substantially larger training budgets.

\subsection{Cross-Domain Generalization}
\label{sec:cross_domain}

To test the generality of \mname{}, we apply it to the Fairy2i method~\cite{wang2025fairy2itrainingcomplexllms}, which transforms a pretrained real-valued model into a lossless, widely-linear complex representation and then performs phase-aware extremely-low-bit quantization in the complex domain~\cite{wang2025ifairy2bitcomplexllm}. This setting provides a complementary testbed where both the computation graph and the quantization codebook differ from standard real-valued QAT, enabling us to assess the robustness and versatility of \mname{} beyond ternary real weights.

Specifically, we apply \mname{} to the Fairy2i pipeline on \texttt{Qwen2.5-0.5B}. We implement the Fairy2i with one recursive residual quantization, yielding a quantization scheme of 2 bits per parameter. 
All variants are trained on 6B tokens from RedPajama under a per-tensor quantization setting (Table~\ref{tab:ifairy_qwen05b}).
Starting from a real-valued BF16 reference, naive real-valued \textsc{Ternary} method causes a clear accuracy drop. In contrast, \textsc{Ternary-H} recovers a substantial portion of the lost performance, and the complex-domain conversion baseline \textsc{Fairy2i} further improves the average accuracy. Finally, our Hessian-guided variant \textsc{Fairy2i-H} achieves the best average score, indicating that \mname{}-style curvature-aware scheduling remains beneficial even when the computation graph and quantization codebook are altered by the complex-valued transformation.

\subsection{Ablation Studies}
\label{sec:ablation}
We conduct targeted ablations to disentangle the effects of (i) quantization granularity (group size) and (ii) Hessian-aware temperature scheduling in \mname{}.

\paragraph{Effect of group size.}
We first ablate the quantization granularity by switching from per-tensor to group-wise quantization with group size 128, while keeping the \textsc{w/o Hessian} training schedule.
As shown in Table~\ref{tab:ablation_groupsize_hessian}, using group size 128 yields a small but consistent improvement in average:
from 0.537 to 0.540 on 1B (+0.56\%) and from 0.597 to 0.599 on 3B (+0.34\%).
This suggests that group-wise quantization provides a modest benefit under our training setup, but does not account for the majority of \mname{}’s gains.

\paragraph{Effect of Hessian-aware Temperature Scheduling.}
Next, we isolate the contribution of curvature guidance by comparing \textsc{w/o Hessian} against full \mname{} under the same group size (128).
On 1B, enabling Hessian-aware temperature scheduling improves the average score from 0.540 to 0.547 (+1.30\%), indicating that curvature-driven scheduling is the primary driver of the gain beyond a uniform annealing baseline.
On 3B, we observe a smaller but still positive improvement from 0.599 to 0.601 (+0.33\%), suggesting that the benefit persists at a larger scale, albeit with reduced magnitude in this budget regime.

\subsection{Computational Efficiency Analysis}
\label{sec:efficiency}
We finally assess the computational and memory overhead of \mname{} relative to standard QAT.
Since \mname{} maintains the same model architecture and utilizes element-wise operations for the soft surrogate, the computational and memory consumption during training are virtually identical to standard STE-based QAT.
The only additional cost stems from the sensitivity estimation via Hutch++ and temperature initialization.
However, as detailed in Sec.~\ref{sec:exp_setup}, these are one-time \emph{offline} pre-computations performed on a small calibration subset.
Empirically, this pre-processing incurs negligible wall-clock time compared to the full training cycle (10B tokens).
Consequently, \mname{} achieves the reported performance gains with almost zero marginal cost over standard QAT baselines.

\section{Conclusion}
We introduce \textbf{\mname{}}, a differentiable QAT framework that resolves the optimization bottlenecks of 1.58-bit LLMs. By replacing the rigid Straight-Through Estimator (STE) with a temperature-controlled \texttt{Softmax} relaxation, \mname{} restores exact gradient fidelity and eliminates the dead zone problem. Furthermore, our Hessian-aware temperature scheduler adaptively modulates the discretization rate based on structural sensitivity, ensuring stable convergence for critical layers. Empirical results of \texttt{LLaMA-3.2} demonstrate that \mname{} significantly narrows the gap between full-precision and extremely low-bit models, proving that prioritizing smooth gradient flow through continuous relaxation is essential for recovering the representational capacity of extremely low-bit architectures.

\clearpage

\bibliography{reference}
\bibliographystyle{icml2026}

\newpage
\appendix
\onecolumn
\section{Proofs for Sec.~\ref{sec:theory}}
\label{app:theory_proofs}

\subsection{Proof of Lemma~\ref{lem:variance_gradient}}
\begin{proof}
Recall $\pi_\tau(q\mid w)$ in Eq.~\eqref{eq:softmax_kernel} and $\mathcal{H}(w;\tau)$ in Eq.~\eqref{eq:relax_op}. Throughout, we treat $\gamma$ as a constant during differentiation as in Sec.~\ref{sec:problem_def}. We prove that
\[
\frac{\partial \mathcal{H}(w;\tau)}{\partial w}
=
\frac{2}{\tau}\,\mathbb{V}_\tau(w),
\]
where $\mu_\tau(w)$ and $\mathbb{V}_\tau(w)$ are defined in Eqs.~\eqref{eq:mu_tau}--\eqref{eq:var_tau}.

\paragraph{Step 1: Reparameterization.}
Let $z\triangleq w/\gamma$. Since $\gamma$ is constant,
\begin{equation}
\frac{\partial z}{\partial w}=\frac{1}{\gamma}.
\label{eq:app_dz_dw}
\end{equation}
Define $a_q(z)\triangleq \exp\!\left(-\frac{(z-q)^2}{\tau}\right)$ and $Z(z)\triangleq \sum_{k\in\mathcal{Q}} a_k(z)$.
Then $\pi_\tau(q\mid w)=\pi_\tau(q\mid z)=a_q(z)/Z(z)$.

\paragraph{Step 2: Derivative of $\pi_\tau(q\mid z)$.}
Differentiate $a_q(z)$:
\begin{equation}
\frac{\partial a_q(z)}{\partial z}
=
-\frac{2(z-q)}{\tau}\,a_q(z).
\label{eq:app_daq}
\end{equation}
Hence
\begin{equation}
\frac{\partial Z(z)}{\partial z}
=
\sum_{k\in\mathcal{Q}} \frac{\partial a_k(z)}{\partial z}
=
-\frac{2}{\tau}\sum_{k\in\mathcal{Q}}(z-k)\,a_k(z).
\label{eq:app_dZ}
\end{equation}
Applying the quotient rule to $\pi_\tau(q\mid z)=a_q(z)/Z(z)$ gives
\begin{align}
\frac{\partial}{\partial z}\pi_\tau(q\mid z)
&=
\frac{(\partial_z a_q)Z-a_q(\partial_z Z)}{Z^2} \nonumber\\
&=
-\frac{2}{\tau}\pi_\tau(q\mid z)(z-q)
+\frac{2}{\tau}\pi_\tau(q\mid z)\sum_{k\in\mathcal{Q}}(z-k)\pi_\tau(k\mid z).
\label{eq:app_dpi_intermediate}
\end{align}
Since $\sum_k \pi_\tau(k\mid z)=1$ and $\mu_\tau(w)=\sum_k k\pi_\tau(k\mid w)=\sum_k k\pi_\tau(k\mid z)$,
\[
\sum_{k\in\mathcal{Q}}(z-k)\pi_\tau(k\mid z)
=
z-\mu_\tau(w).
\]
Substituting into Eq.~\eqref{eq:app_dpi_intermediate} yields
\begin{equation}
\frac{\partial}{\partial z}\pi_\tau(q\mid z)
=
\frac{2}{\tau}\pi_\tau(q\mid z)\bigl(q-\mu_\tau(w)\bigr).
\label{eq:app_dpi_final}
\end{equation}

\paragraph{Step 3: Derivative of $\mu_\tau(w)$.}
By Eq.~\eqref{eq:mu_tau}, $\mu_\tau(w)=\sum_{q\in\mathcal{Q}} q\,\pi_\tau(q\mid z)$, hence
\begin{align}
\frac{\partial \mu_\tau(w)}{\partial z}
&=
\sum_{q\in\mathcal{Q}} q\,\frac{\partial}{\partial z}\pi_\tau(q\mid z) \nonumber\\
&=
\sum_q q\cdot \frac{2}{\tau}\pi_\tau(q\mid z)\bigl(q-\mu_\tau(w)\bigr) \nonumber\\
&=
\frac{2}{\tau}\left(\sum_q q^2\pi_\tau(q\mid z)-\mu_\tau(w)\sum_q q\pi_\tau(q\mid z)\right) \nonumber\\
&=
\frac{2}{\tau}\left(\sum_q q^2\pi_\tau(q\mid w)-\mu_\tau(w)^2\right)
=
\frac{2}{\tau}\,\mathbb{V}_\tau(w),
\label{eq:app_dmu_dz}
\end{align}
where the last equality uses Eq.~\eqref{eq:var_tau}.

\paragraph{Step 4: Conclude $\partial_w \mathcal{H}(w;\tau)$.}
Since $\mathcal{H}(w;\tau)=\gamma\,\mu_\tau(w)$ in Eq.~\eqref{eq:relax_op} and $\gamma$ is constant,
\begin{align}
\frac{\partial \mathcal{H}(w;\tau)}{\partial w}
&=
\gamma\frac{\partial \mu_\tau(w)}{\partial w}
=
\gamma\frac{\partial \mu_\tau(w)}{\partial z}\cdot\frac{\partial z}{\partial w} \nonumber\\
&=
\gamma\cdot \frac{2}{\tau}\mathbb{V}_\tau(w)\cdot \frac{1}{\gamma}
=
\frac{2}{\tau}\,\mathbb{V}_\tau(w),
\end{align}
where we used Eq.~\eqref{eq:app_dz_dw} and Eq.~\eqref{eq:app_dmu_dz}. This proves Lemma~\ref{lem:variance_gradient}.
\end{proof}

\subsection{Proof of Lemma~\ref{lem:boundary_focus}}
\begin{proof}
Recall $\mathcal{H}(w;\tau)$ in Eq.~\eqref{eq:relax_op} and the hard quantizer $Q(w)$ in Eq.~\eqref{eq:hard_quant}, with boundaries $\mathcal{B}=\{-\gamma/2,+\gamma/2\}$ in Eq.~\eqref{eq:true_boundaries}. We prove that, as $\tau\to 0^+$,
\[
\frac{\partial \mathcal{H}(w;\tau)}{\partial w}
\;\xrightarrow{\ \mathcal{D}'\ }\;
\gamma\sum_{b\in\mathcal{B}}\delta(w-b).
\]

\paragraph{Step 1: Piecewise form of $Q(w)$.}
From Eq.~\eqref{eq:hard_quant}, for scalar $w$ let $z=w/\gamma$. Then
$Q(w)=\gamma\cdot\texttt{Clip}(\texttt{Round}(z),-1,+1)$, which implies
\begin{equation}
Q(w)=
\begin{cases}
-\gamma, & w<-\gamma/2,\\
0, & -\gamma/2 < w < \gamma/2,\\
+\gamma, & w>\gamma/2.
\end{cases}
\label{eq:app_Q_piecewise}
\end{equation}
Any additional rounding transitions (e.g., at $|z|=1.5$) do not change the clipped output and hence do not introduce discontinuities.

\paragraph{Step 2: Pointwise limit $\mathcal{H}(w;\tau)\to Q(w)$ for $w\notin\mathcal{B}$.}
Let $z=w/\gamma$ and define $d_q(z)\triangleq(z-q)^2$ for $q\in\mathcal{Q}$.
If $z\notin\{-\tfrac12,+\tfrac12\}$ (equivalently $w\notin\mathcal{B}$), the minimizer
$q^\star(z)\in\arg\min_{q\in\mathcal{Q}} d_q(z)$ is unique.
For any $q\neq q^\star(z)$ define $\Delta_q(z)\triangleq d_q(z)-d_{q^\star(z)}(z)>0$.
Using Eq.~\eqref{eq:softmax_kernel},
\begin{equation}
\frac{\pi_\tau(q\mid w)}{\pi_\tau(q^\star(z)\mid w)}
=
\exp\!\left(-\frac{\Delta_q(z)}{\tau}\right)
\;\xrightarrow[\tau\to 0^+]{}\;0.
\label{eq:app_ratio}
\end{equation}
Since $\sum_{q\in\mathcal{Q}}\pi_\tau(q\mid w)=1$, Eq.~\eqref{eq:app_ratio} implies
$\pi_\tau(q^\star(z)\mid w)\to 1$ and $\pi_\tau(q\mid w)\to 0$ for all $q\neq q^\star(z)$.
Therefore, for all $w\notin\mathcal{B}$,
\begin{equation}
\lim_{\tau\to 0^+}\mathcal{H}(w;\tau)
=
\gamma\sum_{q\in\mathcal{Q}} q\,\lim_{\tau\to 0^+}\pi_\tau(q\mid w)
=
\gamma\,q^\star(w/\gamma).
\label{eq:app_H_pointwise}
\end{equation}
Comparing Eq.~\eqref{eq:app_H_pointwise} with Eq.~\eqref{eq:app_Q_piecewise} yields
$\lim_{\tau\to 0^+}\mathcal{H}(w;\tau)=Q(w)$ for all $w\neq \pm\gamma/2$, hence the convergence holds almost everywhere:
\begin{equation}
\mathcal{H}(w;\tau)\xrightarrow[\tau\to 0^+]{}Q(w)\quad\text{for a.e. }w\in\mathbb{R}.
\label{eq:app_H_to_Q}
\end{equation}

\paragraph{Step 3: Distributional convergence of derivatives.}
Let $\varphi\in C_c^\infty(\mathbb{R})$ be any test function. For each $\tau>0$, $\mathcal{H}(\cdot;\tau)$ is smooth, thus
\begin{equation}
\left\langle \frac{\partial \mathcal{H}(\cdot;\tau)}{\partial w}, \varphi \right\rangle
=
-\left\langle \mathcal{H}(\cdot;\tau), \varphi' \right\rangle
=
-\int_{\mathbb{R}} \mathcal{H}(w;\tau)\,\varphi'(w)\,dw.
\label{eq:app_dist_deriv_def}
\end{equation}
Since $|\mathcal{H}(w;\tau)|\le \gamma$ and $\varphi'$ has compact support,
$|\mathcal{H}(w;\tau)\varphi'(w)|\le \gamma|\varphi'(w)|$ with $\gamma|\varphi'|\in L^1(\mathbb{R})$.
Together with the a.e.\ convergence in Eq.~\eqref{eq:app_H_to_Q}, the dominated convergence theorem yields
\begin{equation}
\lim_{\tau\to 0^+}\int_{\mathbb{R}} \mathcal{H}(w;\tau)\,\varphi'(w)\,dw
=
\int_{\mathbb{R}} Q(w)\,\varphi'(w)\,dw.
\label{eq:app_dct}
\end{equation}
Combining Eqs.~\eqref{eq:app_dist_deriv_def}--\eqref{eq:app_dct},
\begin{equation}
\lim_{\tau\to 0^+}
\left\langle \frac{\partial \mathcal{H}(\cdot;\tau)}{\partial w}, \varphi \right\rangle
=
-\int_{\mathbb{R}} Q(w)\,\varphi'(w)\,dw.
\label{eq:app_reduce_to_Q}
\end{equation}

\paragraph{Step 4: Evaluate the limit functional.}
Using the piecewise form in Eq.~\eqref{eq:app_Q_piecewise},
\begin{align}
\int_{\mathbb{R}} Q(w)\,\varphi'(w)\,dw
&=
\int_{-\infty}^{-\gamma/2} (-\gamma)\,\varphi'(w)\,dw
+
\int_{\gamma/2}^{\infty} (+\gamma)\,\varphi'(w)\,dw \nonumber\\
&=
-\gamma\bigl(\varphi(-\gamma/2)-\varphi(-\infty)\bigr)
+\gamma\bigl(\varphi(\infty)-\varphi(\gamma/2)\bigr).
\label{eq:app_Q_phi_prime}
\end{align}
Because $\varphi$ has compact support, $\varphi(\pm\infty)=0$, hence
\begin{equation}
\int_{\mathbb{R}} Q(w)\,\varphi'(w)\,dw
=
-\gamma\,\varphi\!\left(-\frac{\gamma}{2}\right)
-\gamma\,\varphi\!\left(+\frac{\gamma}{2}\right).
\label{eq:app_Q_phi_prime_simplify}
\end{equation}
Plugging Eq.~\eqref{eq:app_Q_phi_prime_simplify} into Eq.~\eqref{eq:app_reduce_to_Q} yields
\[
\lim_{\tau\to 0^+}
\left\langle \frac{\partial \mathcal{H}(\cdot;\tau)}{\partial w}, \varphi \right\rangle
=
\gamma\,\varphi\!\left(-\frac{\gamma}{2}\right)
+\gamma\,\varphi\!\left(+\frac{\gamma}{2}\right)
=
\left\langle \gamma\delta\!\left(w+\frac{\gamma}{2}\right)+\gamma\delta\!\left(w-\frac{\gamma}{2}\right),\,\varphi(w)\right\rangle.
\]
Since $\varphi$ is arbitrary, we conclude
\[
\frac{\partial \mathcal{H}(w;\tau)}{\partial w}
\;\xrightarrow[\tau\to 0^+]{}\;
\gamma\sum_{b\in\mathcal{B}}\delta(w-b),
\]
which is Eq.~\eqref{eq:dirac_limit} and completes the proof of Lemma~\ref{lem:boundary_focus}.
\end{proof}

\section{Implementation Details}
\label{app:imp_details}

We provide the detailed hyperparameter configurations used for \mname{} training in Table~\ref{tab:hyperparams}.

\begin{table}[h]
\centering
\caption{Hyperparameter settings for \mname{} quantization-aware training.}
\label{tab:hyperparams}
\begin{tabular}{l c}
\toprule
\textbf{Hyperparameter} & \textbf{Value} \\
\midrule
\multicolumn{2}{l}{\textit{General Training Settings}} \\
\midrule
Global Batch Size & 256 \\
Optimizer & AdamW \\
Learning Rate & 1.5e-3 \\
LR Scheduler & WSD \\
Weight Decay & 0.1 \\
\midrule
\multicolumn{2}{l}{\textit{\mname{} Specifics}} \\
\midrule
Pressure Ratio $\rho$ & 0.2 \\
Temperature Scaling Strength $\alpha$ & 0.4 \\
Initial Temperature & 0.3 \\
Temperature Schedule & Cosine \\
Group Size & 128 \\
\midrule
\multicolumn{2}{l}{\textit{Sensitivity Estimation (Hutch++)}} \\
\midrule
Calibration Batch Size & 50 \\
Sketch Rank $r$ & 10 \\
Hutchinson Samples & 20 \\
Noise Type & Rademacher \\
Sensitivity Gain $\kappa$ & 1.0 \\
Normalization Epsilon $\epsilon$ & 1e-8 \\
\bottomrule
\end{tabular}
\end{table}

\end{document}